\documentclass[runningheads]{llncs}
\usepackage{graphicx}
\usepackage{amsmath} 
\usepackage{caption}
\usepackage{subcaption}
\usepackage{hyperref}

\RequirePackage[
	style=numeric,
	sorting=none,
	language=autobib,
	autolang=other,
	urldate=iso8601,
	backref=false,
	isbn=true,
	url=false,
	maxbibnames=3,
	backend=bibtex  
]{biblatex}

\RequirePackage{csquotes}
\begin{document}
%
\title{Real-time FPGA implementation of the Semi-Global Matching stereo vision algorithm for a~4K/UHD video stream}
\titlerunning{Real-time FPGA implementation of the SGM stereo vision in 4K}
%
\author{Mariusz Grabowski \href{https://orcid.org/0000-0002-6686-7390}{\includegraphics[width=16pt]{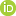}} \and
Tomasz~Kryjak \href{https://orcid.org/0000-0001-6798-4444}{\includegraphics[width=16pt]{figures/orcid.png}}}

\authorrunning{M. Grabowski et al.}
%
\institute{Embedded Vision Systems Group, Computer Vision Laboratory, \\ Department of Automatic Control and Robotics, \\ AGH University of Science and Technology, Krakow, Poland
\email{grabowski@student.agh.edu.pl, tomasz.kryjak@agh.edu.pl}} 
\maketitle              
\begin{abstract}

In this paper, we propose a~real-time FPGA implementation of the Semi-Global Matching (SGM) stereo vision algorithm.
The designed module supports a~4K/Ultra HD ($3840 \times 2160$ pixels @ 30 frames per second) video stream in a~4 pixel per clock (ppc) format and a~64-pixel disparity range.
The baseline SGM implementation had to be modified to process pixels in the 4ppc format and meet the timing constrains, however, our version provides results comparable to the original design.
The solution has been positively evaluated on the Xilinx VC707 development board with a~Virtex-7 FPGA device.




\end{abstract}
%
%
%
\section{Introduction}

Information on the 3D structure (depth) of a~scene is very important in many robotic systems, including self-driving cars and unmanned aerial vehicles (UAVs), as it is used in object detection and navigation modules.
The depth map can be estimated using several different approaches, active: LiDAR (Light Detection and Ranging), Time of Flight (ToF) cameras, stereo vision with structured lighting; and passive: stereo vision. 
Stereo vision uses two or more cameras that acquire the same scene, but from slightly different points in space.
A~detailed discussion of the advantages and disadvantages of different sensors can be found in the work of Jamwal, Jindal, and Singh \cite{Jamwal_2016}.



Stereo vision, in its passive variant, is an often used solution in embedded systems due to the low price of the equipment, its small size and weight (no need for a~laser light source, rotating elements or projectors). 
The accuracy of the results obtained with this technology strictly depends on the algorithm used to process the acquired images. 
The methods used can be divided into two groups: local and~global \cite{Scharstein_2002}. 
In both cases, the key element is to find the same pixels in the image captured by the left (usually considered as the base) and right camera (reference).
Their offset expressed in pixels is referred to as the disparity.
This value can be easily converted to the distance from the sensors using the vision system parameters.

Global methods introduce appropriate discontinuity penalties in~order to smooth the disparity map. 
Their aim is to optimise the energy function defined over the whole image. 
By means of global algorithms, much more reliable and~accurate disparity maps are determined, but the smoothing task is NP-hard and~algorithms are very computationally demanding, and~for this reason they are not suitable for implementation in real-time systems.

It should be also noted that the current dominant trend is depth estimation using deep neural networks \cite{Laga_2022}. However, due to the high computational complexity, especially for high-resolution video streams, this topic remains outside the focus of our present work.



The SGM (Semi-Global Matching) algorithm was introduced by Hirshmüller in 2005 \cite{Hirschmuller_2005} and 2008 \cite{Hirschmuller_2008}.
It is based on two components: (1) matching at a~single pixel level with the~use of mutual information and (2) approximation of a~global, two-dimensional smoothness constraint (obtained by combining multiple 1D constraints).
The SGM algorithm is an example of an intermediate method between local and~global approaches for determining disparity maps and is a~compromise between accuracy and~computational complexity.
However, using SGM for high-resolution images is still challenging. For example, for a~resolution of $1920 \times 1080$ pixels at 30 frames per second, an execution of about 2 TOPS (Tera Operations Per Second) with memory bandwidth of 39 Tb/s is required to process all pixels (2 million) \cite{Lee2021}.


In this paper we present an architecture of a~stereo vision system with a~modified SGM algorithm to process a~4K/Ultra HD ($3840 \times 2160$ pixels @ 30 frames per second) video stream in 4ppc (pixel per clock) format and its implementation in an FPGA (Field Programmable Gate Array) device. 
The proposed modification solves the data dependency problem while not affecting the algorithm's accuracy.
To the authors' knowledge, this is the only verified hardware implementation of the SGM method for 4K/Ultra HD resolution.


The reminder of this paper is organised as follows.
In Section \ref{sec:sgm_algorithm} we present basic information about the SGM algorithm. 
In Section \ref{sec:prev_work} we review the previous work on SGM implementation on FPGAs.
We describe the proposed method and architecture, as well as the evaluation of the algorithm and the hardware implementation in Section \ref{sec:arch}.
The paper ends with conclusions and future research directions.


\section{The SGM algorithm}
\label{sec:sgm_algorithm}

As mentioned in the introduction, the SGM algorithm is an intermediate approach between local and global methods for determining disparity maps.
Furthermore, it is possible to implement it in an FPGA, in a~pipelined vision system.



The input to the algorithm is a~pair of rectified images.
It consists of the following steps: calculation of the matching cost, aggregation of the cost (calculation of the smoothness constraint) and determination of the final disparity map.





\begin{figure}[!t]
\centering
\includegraphics[width=0.7\textwidth]{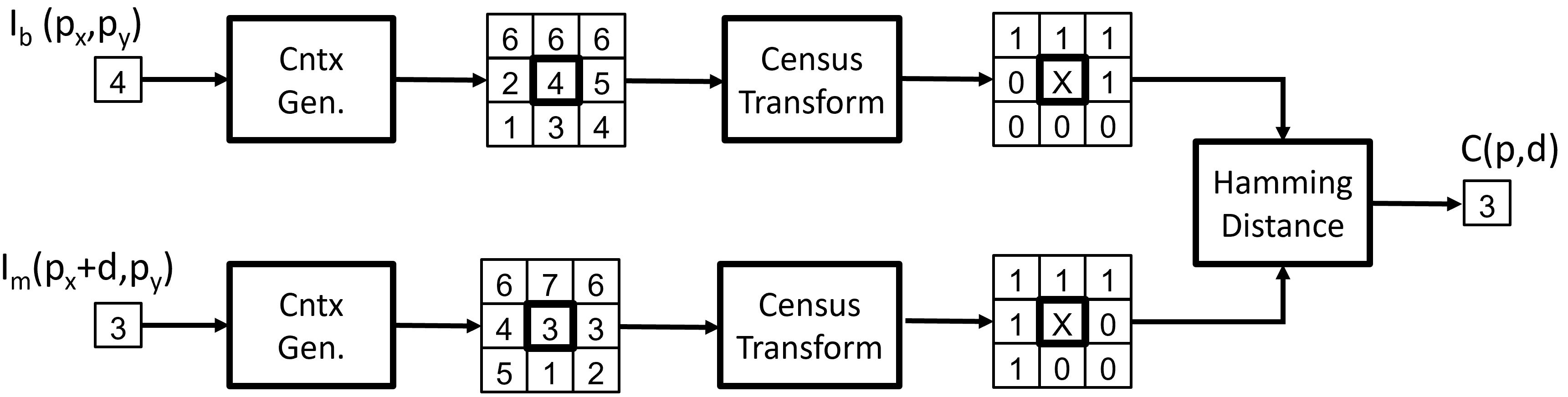}
\caption{\label{fig:cesnus} Matching cost calculation with the Census transform and the Hamming distance metric, with example values.} 
\end{figure}

In this work, the cost of matching $C(p,d)$ between a~pixel $\mathbf{p}=[p_x,p_y]^T$ from the base image $I_b$, and the potentially corresponding pixel (shifted by the disparity $d$ in a horizontal line) in the reference image $I_m$, is calculated using the Census transform and the Hamming distance measure, as shown in Figure \ref{fig:cesnus}.


Determining the correspondence between pixels using only the matching cost alone can lead to ambiguous and incorrect results.
Therefore, an additional global condition is proposed in the SGM algorithm, which adds a~``penalty'' for changing the disparity value (i.e, supports the smoothness of the image), by aggregating the costs along independent paths.


Let $L_\mathbf{r}$ denote the path in the direction $\mathbf{r}$.
The path cost $L_\mathbf{r}(\mathbf{p},d)$ is defined recursively as:
\begin{equation}
\label{eq:pathcost} 
\begin{aligned}
L_{r}(\textbf{p}, d) = C(\textbf{p}, d) + min[L_{\textbf{r}}(\textbf{p} - \textbf{r}, d), \\
L_{\textbf{r}}(\textbf{p} - \textbf{r}, d - 1) + P_{1},\\ L_{\textbf{r}}(\textbf{p} - \textbf{r}, d + 1) + P_{1},\\ \min\limits_{i} L_{\textbf{r}}(\textbf{p} - \textbf{r}, i) + P_{2}]\\  - \min\limits_{k} L_{\textbf{r}}(\textbf{p} - \textbf{r}, k)
\end{aligned}
\end{equation}
where: $C(p,d)$ is the matching cost, and the second part of the equation is the minimum path cost for the previous pixel $\mathbf{p}-\mathbf{r}$ on the path, taking into account the corresponding discontinuity penalty. Two penalties were applied in the~algorithm, $P1$ for a~1-level change in disparity and $P2$ for a~larger change.


Finally, the matching cost is given as:
\begin{equation}
\label{eq:aggrcost}
S(\textbf{p}, d) = \displaystyle\sum_{\textbf{r}} L_{\textbf{r}}(\textbf{p}, d)
\end{equation}  

The author of SGM recommend aggregation along at least 8 paths, i.e, vertically, horizontally and diagonally in both directions (cf. Figure \ref{fig:path_direction}), although he suggests that good results are achieved for the number 16.
The disparity map $D_b$ corresponding to the base image $I_b$ is obtained by selecting for each value $\mathbf{p}$ the disparity $d$ that corresponds to the minimum cost i.e, $min_d S(\mathbf{p},d)$.
Optional element of the algorithm is the final post-processing: median filtering and map consistency check (so called left-right consistency check).


Due to the reasonable trade-off between computational complexity and the quality of the resulting disparity map, the SGM algorithm has become very popular.
It is a~basic method in the popular OpenCV library and the Computer Vision Toolbox of the Matlab software.
It also provides an attractive solution for hardware implementations in FPGAs.



\section{Previous work}
\label{sec:prev_work}

The topic of implementing stereo correspondence using FPGAs is very extensive, and hence this review is narrowed only to selected articles describing the SGM algorithm.
Interested readers are referred to the review \cite{Wan_2021}.


The paper written by Gehrig, Eberli, and Meye in 2009 \cite{Gehrig_2009} described an SGM architecture for processing images with a~resolution of $750 \times 480$ pixels (effectively $340 \times 200$) @ 27 fps at 64 levels of disparity. It is worth noting that this was the first implementation of the SGM method in an FPGA. 


The paper of Hofmann, Korinth, and Koch from~2016 \cite{Hofmann_2016} also proposes a~hardware implementation of the SGM algorithm.
The architecture features scalability and combines coarse-grain and fine-grain parallelisation capabilities.
The authors performed tests for different configurations and resolutions.
For $1920 \times 1080$ pixels @ 30 fps and 128 disparity levels, real-time processing was achieved at a~clock of 130 MHz (VC709 board with Virtex-7 FPGA device).


In the paper of Zhao et al. from 2020 \cite{Zhao_2020}, the authors presented the FP-Stereo library, which uses the HLS language and allows the creation of SGM disparity calculation modules.
The module has been designed in the form of an accelerator interfacing with a~DMA controller, rather than directly with the video stream.
For a~300 MHz clock, a~resolution of $1242 \times 374$ pixels and 128 disparity range, 161 fps were achieved on the ZCU 102 board with the Xilinx Zynq UltraScale+ MPSoC device.


In~the~latest~publications by Shrivastava et al. in~2020~\cite{Shrivastava2020} and Lee with Kim in~2021~\cite{Lee2021}, the support for parallel pixel processing has been added to increase throughput.
In this approach, the main challenge is the presence of an inherent data dependency. 
In the paper from 2020 \cite{Shrivastava2020}, it is addressed by \textit{dependency relaxation}, i.e, the aggregation is performed on the basis of the pixel \(k\)~earlier, where \(k\)~is the number of pixels processed simultaneously. 
The author points out that such a~solution represents a~trade-off between accuracy and~throughput. 

In the work from 2021 \cite{Lee2021}, on the other hand, a~different approach is presented, in~which operations involving the inherent data dependency are performed not on a~single pixel, but on a~vector of pixels.
This allows the generation of disparity maps with~very close accuracy to the original SGM algorithm. 
In~both solutions, the matching costs are determined based on the Census transform.   
In the first publication \cite{Shrivastava2020}, for images at a~resolution of $1280 \times 960$ pixels and~disparity range of 64, 322~fps, and~in~the~second \cite{Lee2021} for a~resolution of $1920 \times 1080$ pixels and~disparity range of 128, 103~fps were obtained.



We also propose a~solution to the inherent data dependency problem. 
Our architecture is based on estimating the previous pixel aggregation cost on a path with minimal additional logic needed. That allows us to process images with a~4K resolution and also to obtain comparable results to the original SGM algorithm without parallelism. 
\section{The proposed hardware implementation}

\label{sec:arch}

\begin{figure}[!t]
\centering
\includegraphics[width=0.8\textwidth]{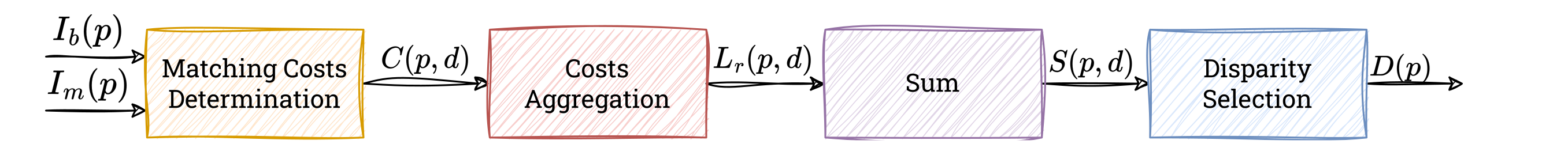}
\caption{\label{fig:schemat_full_sgm} A~general scheme of the proposed SGM disparity estimation system.} 
\end{figure}

The aim of our work was to implement a~hardware architecture capable of processing a~video stream with a~resolution of $3840 \times 2160$ pixels in real-time (i.e~processing 30~frames per second with no pixel dropping).
That stream transmitted in a~1 pixel per clock format requires a~pixel clock frequency of approximately 250~MHz. 
Adding to this value i.e, the vertical and~horizontal blanking fields, the required clock equals about 300~MHz, which is too high for the rather complicated SGM algorithm. At the bottleneck, cost aggregation calculations take more than 10 ns on our platform. 
So, in order to process the data in the desired resolution, it is necessary to introduce parallelisation. 
In this work, a~4ppc (pixel per clock) format is used, in which 4 pixels are processed in parallel.
This allows the pixel clock to be lowered to approximately 75 MHz.
However, the use of such format has significant implications on the implementation of the SGM algorithm, due to the inherent data dependency.

A~general scheme for the proposed vision system is shown in Figure \ref{fig:schemat_full_sgm}.
The module accepts a~synchronised video stream of rectified images, the base \(I_B(p)\) and the reference \(I_M(p)\) one. 
Further processing consists of several steps: determination of the matching cost \(C(p,d)\) using the Census transform based matching method, calculation of the cost aggregation \(L_r(p,d)\), summation of the aggregation costs from all directions \(S(p,d)\) and disparity determination \(D(p)\).



\subsection{Determination of the matching cost}

The 4ppc format does not introduce major complications into the hardware architecture of the matching cost determination module, but only increases the hardware resource requirements.
First, $5 \times 5$ contexts are created for both images. 
For the base image, in a~given cycle, 4~contexts are created (as implied by the 4ppc format \cite{Kowalczyk2018}), and for the reference image this number is increased by the disparity range ($4 + disp\_range - 1$), so that it is possible to simultaneously compare each of the 4 contexts of the base image with all the contexts in the disparity range of the reference image.
A~Census transform is performed on the generated contexts, and the contexts are then compared accordingly using the Hamming distance metric. 
The output consists of matching cost vectors.




\subsection{Cost aggregation}

\begin{figure}[!t]
\centering
\includegraphics[width=0.4\textwidth]{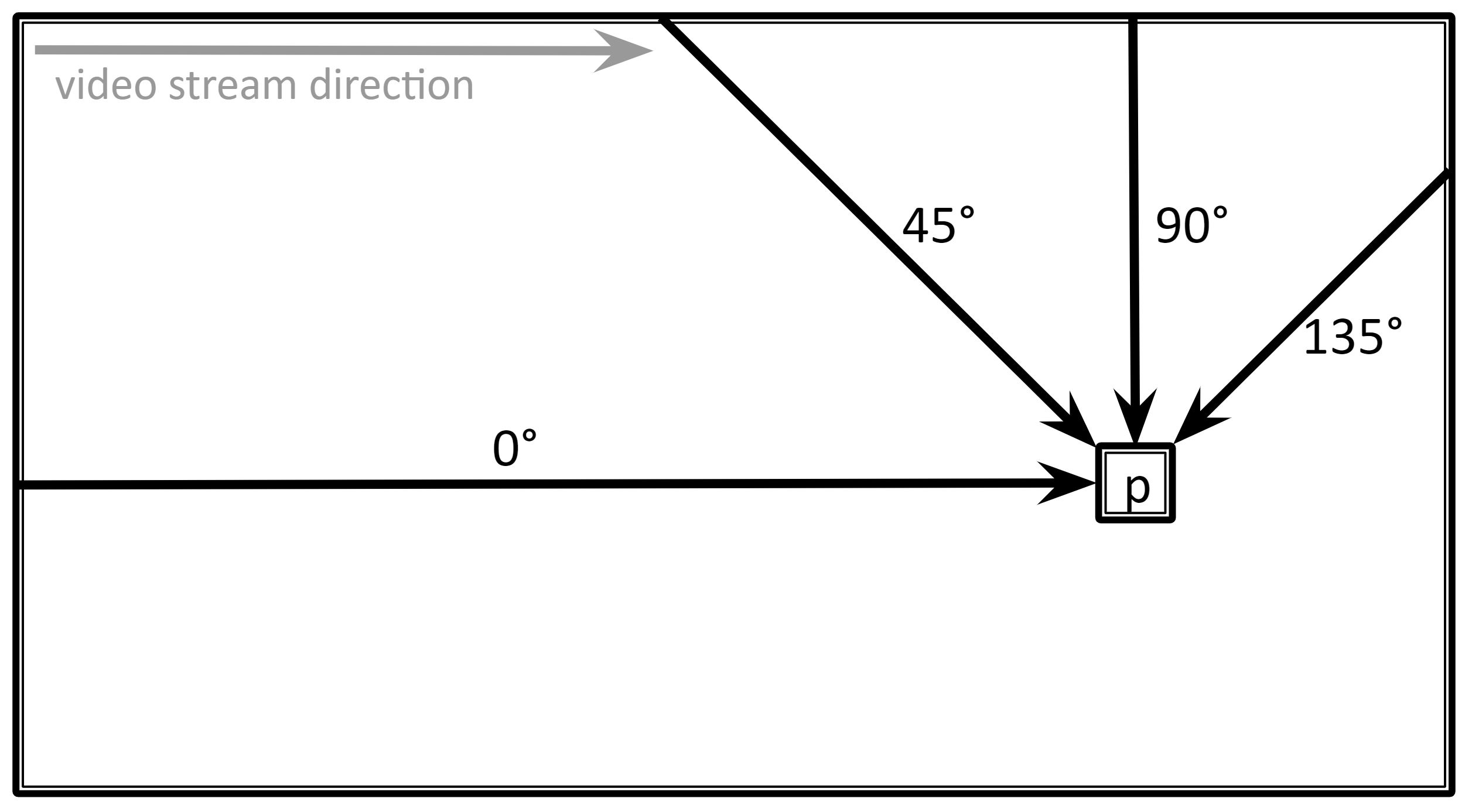}
\caption{\label{fig:path_direction} Cost aggregation paths in SGM.}
\end{figure}

In the next step, a~quasi-global optimisation is performed by aggregating the costs for the whole image according to the SGM algorithm. 
In the current version of the module, this is implemented on four paths in the directions 0°, 45°, 90°, 135°, as shown in Figure \ref{fig:path_direction}, which can be processed directly (without additional video stream buffering). 


Theoretically, it is also possible to realise the other four directions (180°, 225°, 270°, 315°), but this would require storing the entire image in external RAM, using additional resources of the FPGA device, complex control logic and introducing additional latency in image processing. 

\begin{figure}[!t]
\centering
\includegraphics[width=0.9\textwidth]{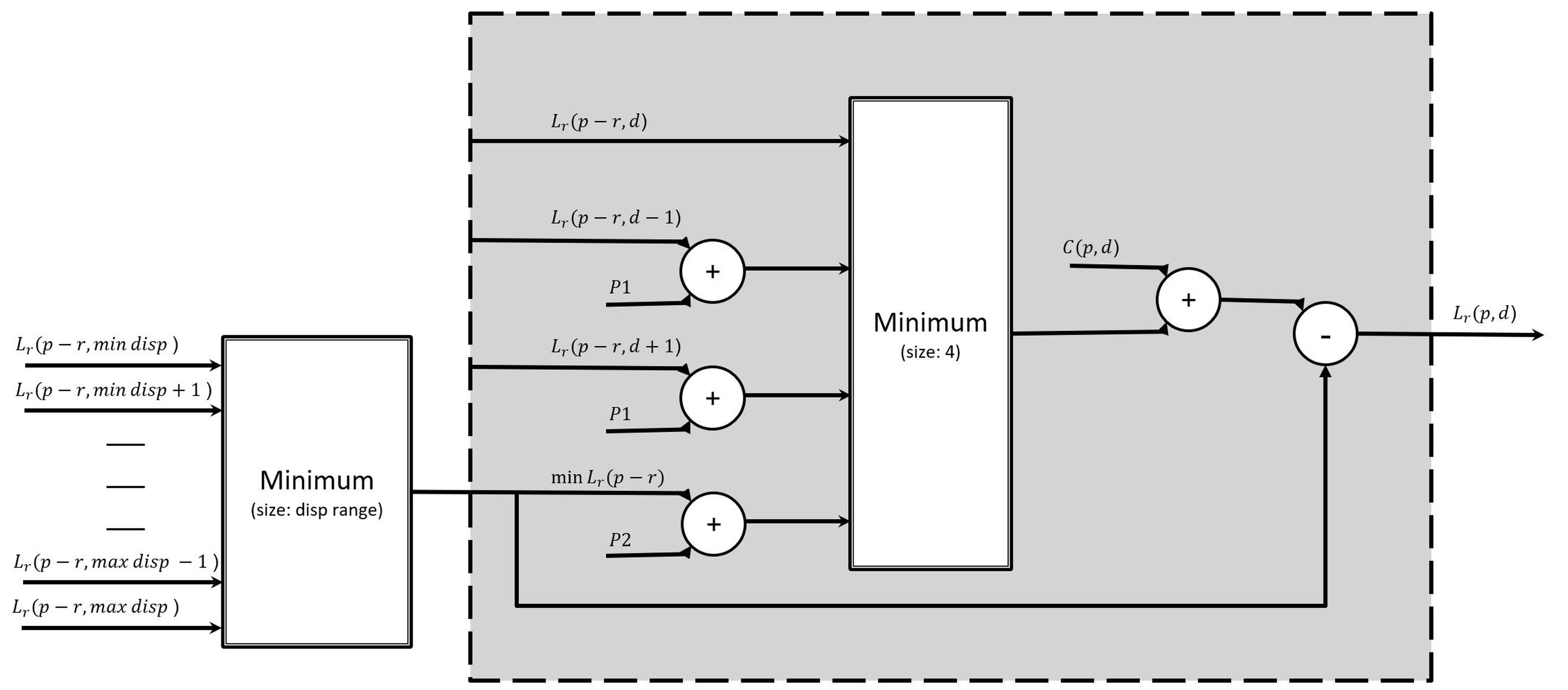}
\caption{\label{fig:l_calculations} Hardware architecture of the aggregation cost calculation unit for path \(r\), pixel \(p\) and disparity \(d\).}
\end{figure}

In order to calculate the aggregation cost for a~given pixel, it is necessary to know the value of the aggregation cost for the previous pixel on the path (cf. Equations \eqref{eq:pathcost} and \eqref{eq:aggrcost}). 
For the 45°, 90°, 135° paths, the aggregation costs for the pixels in a~given line are stored in Block RAM and read out accordingly during the processing of the next image line to calculate the costs for the subsequent pixels on these paths. 
The hardware architecture of this computation is shown in Figure \ref{fig:l_calculations} and follows Equation \eqref{eq:pathcost}. 
The grey part is replicated for the entire range of disparities (\emph{disp range}) and performs in parallel and one block of finding the minimum value of aggregation costs of the previous pixel on the path \(min L_r(p-r)\) is exploited to calculate the aggregation cost for the current pixel for each disparity value in the range.


For the 4ppc format, the difficulty arises for the 0° path. 
Using the aggregation cost of the previous pixel  \(L_{{r}}({p} - \textbf{r}, d)\), which for this path lies in the same image line and potentially in the same 4ppc format data vector, results in the need to process four pixels in the same clock cycle.
In the worst case, for the last pixel in the vector, in one clock cycle the data would have to propagate through four serially connected aggregation cost calculation units, as in Figure \ref{fig:l_calculations}. 
The critical path would contain 4 minimum modules of size \emph{disp range}, four minimum modules of size 4 and 12 adders/subtractors. 
For this reason, the cost aggregation based on a~baseline architecture (i.e, as proposed by the authors of SGM) for the 0° path is not feasible for the considered 4K resolution, without violating timing constraints.


It is therefore necessary to propose a~new solution for the calculation of the aggregation cost for the 0° path. 
Time constraints require that the new architecture does not introduce significant additional propagation time and maintains the approximation assumption of the global smoothness constraint of the SGM algorithm.


In our work, we designed and implemented an architecture with a~proposed estimation of the aggregation cost value for consecutive pixels based on the calculated aggregation cost for the last pixel of the previous 4ppc vector (the pixel processed in the previous clock cycle) and the matching costs of the previous pixels in the same 4ppc vector.

    For the first pixel in the 4ppc vector, the aggregation cost of the previous pixel is available during the calculation (it was calculated for the previous 4ppc vector), i.e:
    
    \begin{equation}
        L_r(p_1 - r, d) = L_r(p_{last}, d)
    \end{equation}
    
    \noindent where: \(L_r(p_1 - r, d)\) is the aggregation cost of the previous pixel relative to the first pixel in the 4ppc vector (\(p_1 - r \)), and \(L_r(p_{last} - r, d)\) is the aggregation cost of the last pixel in the previous 4ppc vector.
    
\begin{figure}[!t]
\centering
\includegraphics[width=0.5\textwidth]{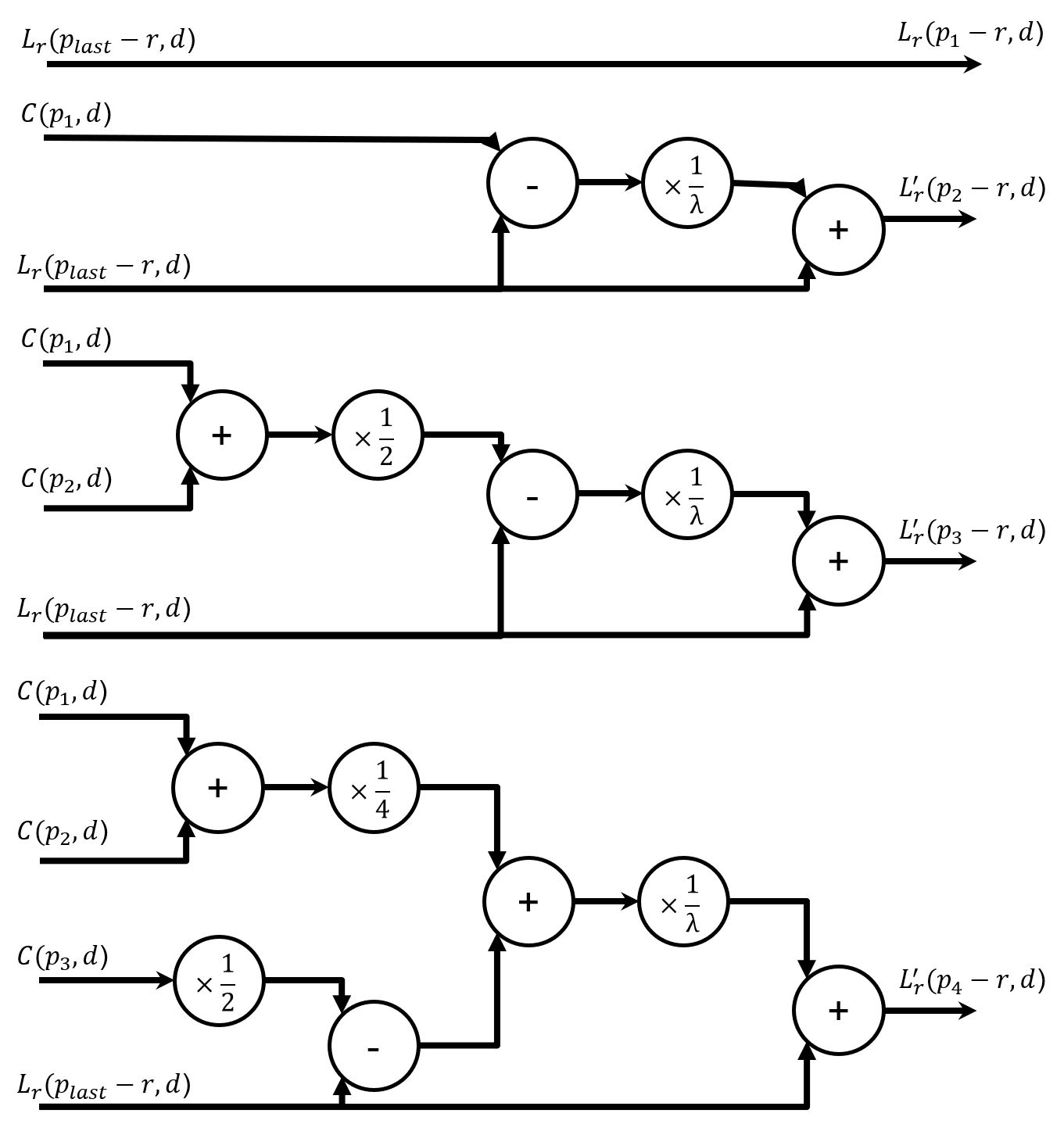}
\caption{\label{fig:l_estimation} The architecture for estimating the aggregation cost of the previous pixel for each pixel in the 4ppc vector.}
\end{figure}

    
    For the consecutive pixels, we propose an estimation, which is performed according to the following Equations:
    
    
    \begin{equation}
    \begin{aligned}
      & L'_{r}(p_2 - r, d) = L_r(p_{last}, d) + \dfrac{1}{\lambda}(C(p_1, d)-L_r(p_{last}, d)) \\
      & L'_{r}(p_3 - r, d) = L_r(p_{last}, d) + \dfrac{1}{\lambda}(\dfrac{C(p_1, d)+C(p_2, d)}{2}-L_r(p_{last}, d)) \\
      & L'_{r}(p_4 - r, d) = L_r(p_{last}, d) + \dfrac{1}{\lambda}(\dfrac{\dfrac{C(p_1, d)+C(p_2, d)}{2}+C(p_3, d)}{2}-L_r(p_{last}, d)) 
    \end{aligned}
    \end{equation}
   \noindent  where: \(L'_{r}(p - r,d)\) is the estimated aggregation cost for the previous pixel relative to the pixel \(p\), \(C(p, d)\) is the matching cost for a~given pixel, and the coefficient \(\lambda\) may take a~value which is a~power of two (\(1,2,4,8,16,...\)). The architecture of this solution is shown in Figure \ref{fig:l_estimation}.


The algorithm is based on the difference of the matching cost values of the previous pixels in a~given 4ppc vector with the aggregation cost for the last pixel of the previous vector. 
The aggregation cost estimation architecture consists of basic components and introduces an additional delay only by the propagation time of the 3 adders/subtractors (critical path for \(L_{r}(p_4 - r, d)\). 
Note: multiplication/division by a~number that is a~power of two is only a~bit shift and requires no delay in the hardware implementation. 
 

The solution takes into account the matching cost values of all previous pixels with the possibility to adjust the impact of the matching cost of previous pixels in a~given vector by a~factor of \(\lambda\).


The estimated aggregation costs are then used to calculate the aggregation costs according to the architecture in Figure \ref{fig:l_calculations}.
In the work of Shrivastava et al. \cite{Shrivastava2020} the estimation has been omitted and in the work of Lee and Kim \cite{Lee2021} it has been solved by the cluster-wise cost aggregation.

The aggregation costs from all paths are then summed and the disparity is calculated. 
This involves finding the minimum matching cost.


\begin{figure}[!t]
	\centering
	\begin{subfigure}[t]{0.29\textwidth}
	     \includegraphics[width=\textwidth]{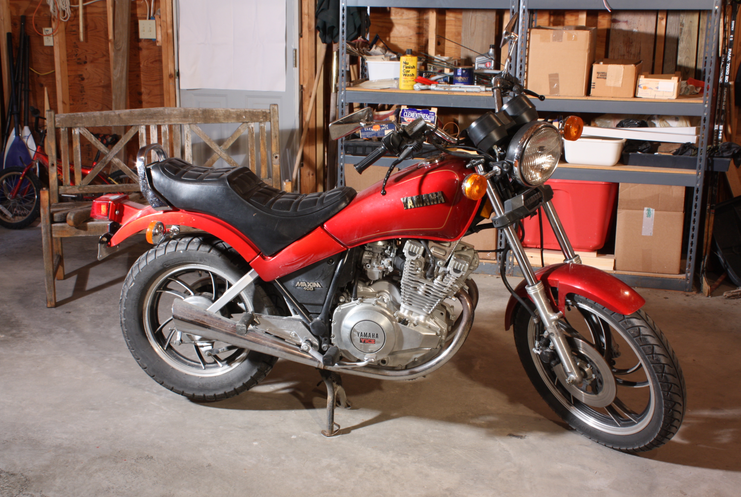}
         \caption{Input image -- left}
    \end{subfigure}
     ~
	\begin{subfigure}[t]{0.29\textwidth}
         \includegraphics[width=\textwidth]{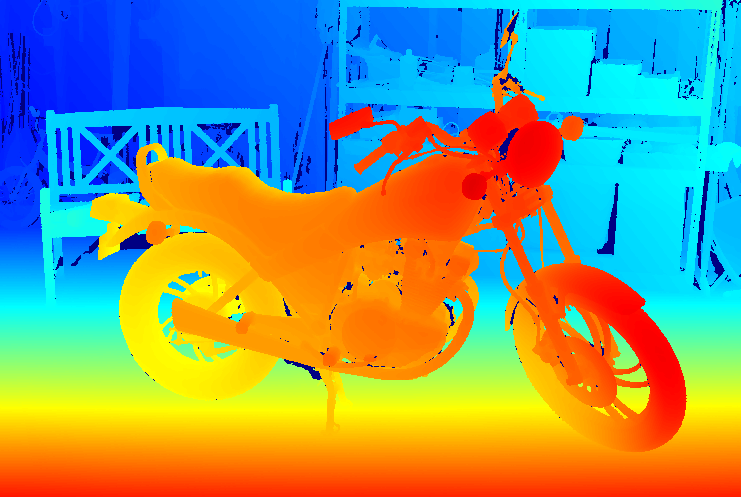}
         \caption{Ground truth}
    \end{subfigure}
    \\
    \begin{subfigure}[t]{0.29\textwidth}
	     \includegraphics[width=\textwidth]{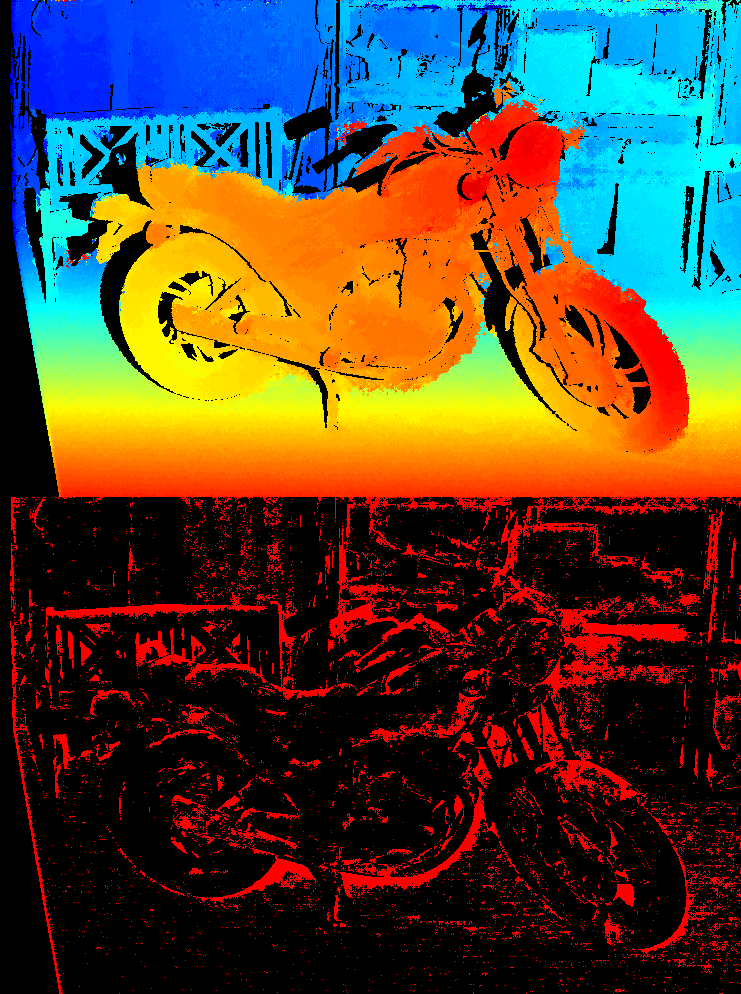}
         \caption{SGM 4ppc}
    \end{subfigure}
     ~
    \begin{subfigure}[t]{0.29\textwidth}
	     \includegraphics[width=\textwidth]{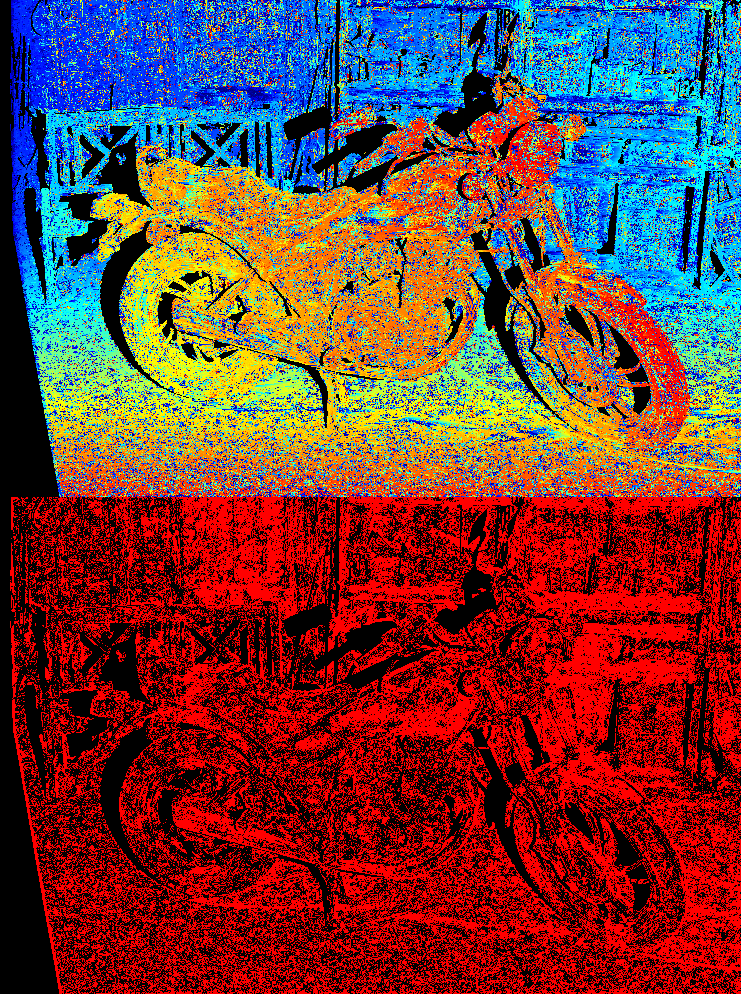}
         \caption{Local method based on CT}
    \end{subfigure}
    \\
    \begin{subfigure}[t]{0.29\textwidth}
	     \includegraphics[width=\textwidth]{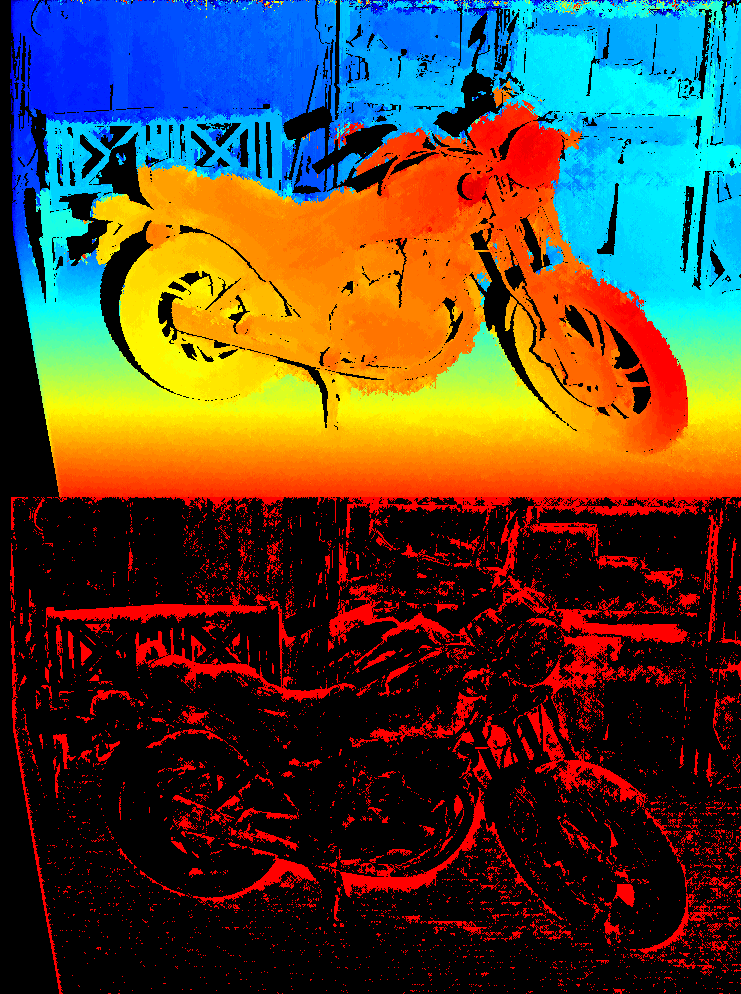}
         \caption{SGM -- 3 paths }
    \end{subfigure}
     ~
    \begin{subfigure}[t]{0.29\textwidth}
	     \includegraphics[width=\textwidth]{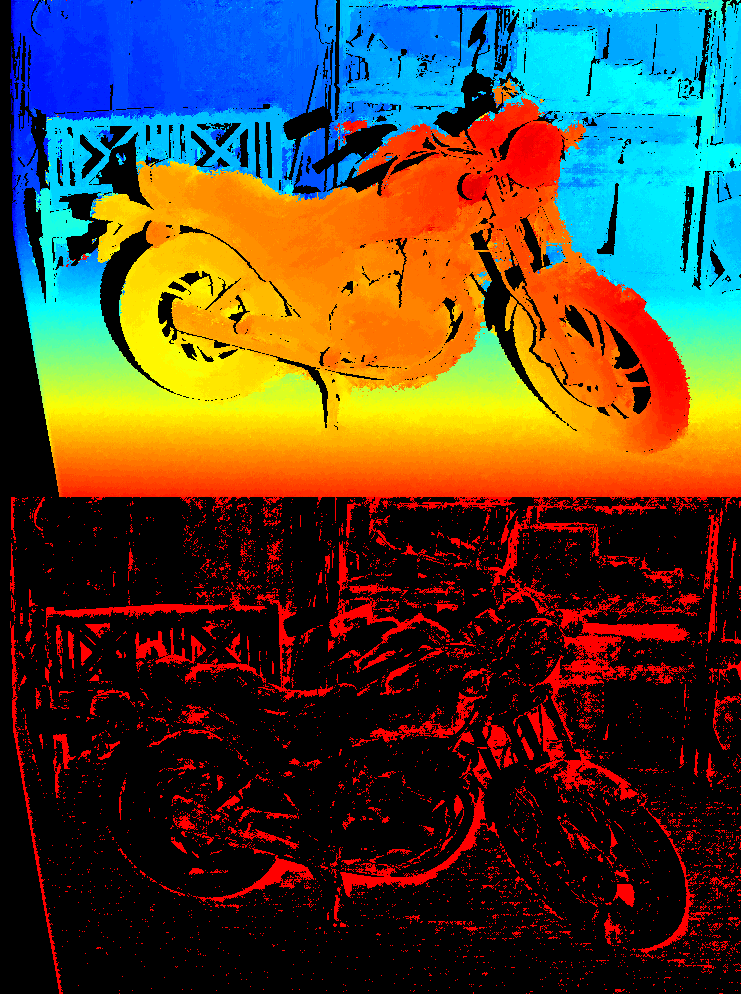}
         \caption{SGM -- 4 paths }
    \end{subfigure}
     ~
    \caption{Comparison of output disparity maps for the \textit{Motorcycle} image in Middlebury 2014 dataset: (a) the left input image, (b) the ground truth disparity map, (c), (d), (e), (f) estimated disparity maps (on the top) and the error maps (on the bottom).}%
	\label{fig:motorcycle}%
\end{figure}


\subsection{Evaluation of the proposed method}

The accuracy evaluation of the proposed algorithm was performed on a~set of stereo images from the Middlebury 2014 \cite{Scharstein2014} dataset.
We skipped the final post-processing to better highlight the differences between the base SGM algorithm and the modified version proposed in this paper (SGM 4ppc).
The accuracy was measured by the ratio of pixels with incorrect disparity value to all pixels of the image (all) and also to the non-occluded (noc) pixels (occluded pixels should be filled with the Left/Right Check post-processing).


We compared the proposed method (SGM 4ppc) with the conventional local block matching based on the Census transform and the SGM algorithm (also based on the Census transform) with 3 and 4 aggregation paths.
Figure \ref{fig:motorcycle} shows sample evaluation results on the \textit{Motorcycle} images from the Middlebury 2014 dataset. Table \ref{tab1} shows the average evaluation results for the entire dataset. 

The accuracy of the proposed method is comparable to the original SGM algorithm with 4 paths. The difference between error rates is about 0.4\%.

\begin{table}[!t]
\centering
\caption{Comparison of error rates for the Middlebury 2014 dataset, based on all (all) and non-occluded (noc) pixels.}\label{tab1}
\begin{tabular}{p{0.25\textwidth} p{0.25\textwidth} p{0.25\textwidth}}
\hline
 & \multicolumn{1}{c}{\textbf{all}} & \multicolumn{1}{c}{\textbf{noc}} \\
\hline
Local based on CT & \multicolumn{1}{c}{68.21\%} & \multicolumn{1}{c}{63,36\%} \\
\hline
SGM 3 paths & \multicolumn{1}{c}{38.01\%} & \multicolumn{1}{c}{28.79\%} \\
\hline
SGM 4 paths & \multicolumn{1}{c}{36.27\%} & \multicolumn{1}{c}{26.88\%} \\
\hline
SGM 8 paths & \multicolumn{1}{c}{33.31\%} & \multicolumn{1}{c}{23.11\%} \\
\hline
\textbf{SGM 4ppc} & \multicolumn{1}{c}{\textbf{36.64}\%} & \multicolumn{1}{c}{\textbf{27.32}\%} \\
\hline
& & \\
\end{tabular}
\end{table}

\subsection{Hardware implementation}

We implemented the proposed stereo vision system on a~VC707 evaluation board with Xilinx's Virtex-7 XC7VX485T-2FFG1761C device. 
We set up a~test environment to evaluate the system, with test images sent directly from a~PC do the board and later displayed on a~4K monitor. 


We compared our solution with previous FPGA implementations of the SGM algorithm in Table \ref{tab2}.
We used the following metrics: Frames per Second (FPS), Million Disparity Estimates per second (MDE/s) and MDE/s per Kilo LUTs (Look-Up Tables) (MDE/s/KLUT).
First of all, our solution is the only one verified in hardware for a~4K/ Ultra HD resolution.
We also would like to point out that the lower performance in FPS and MDE/s relative to previous work from 2020 \cite{Shrivastava2020} and 2021 \cite{Lee2021} is due to the use of an FPGA chip with fewer resources. 
For this work, it was necessary to select a~suitable platform to enable image acquisition in 4K resolution (i.e, having two high-bandwidth FMCs (FPGA Mezzanine Connectors) to which TB-FMCH-HDMI4K modules were attached).

It is also worth mentioning that the used FPGA technology differs not only in the number of resources but also in the performance.
To compare: the critical path propagation time for the technology used in this paper after synthesis is 12.967 ns, but for the Xilinx Virtex UltraScale+ XCVU9P-L2FLGA2104E FPGA with the same parameters, it is 8.240 ns (36.45\% faster).


\begin{table}[!t]
\centering
\caption{Comparison with previous FPGA implementations of the SGM algorithm.}\label{tab2}
\begin{tabular}{p{0.05\textwidth}  p{0.10\textwidth}  p{0.08\textwidth}  p{0.10\textwidth}  p{0.05\textwidth} p{0.05\textwidth} p{0.05\textwidth}  p{0.05\textwidth} p{0.05\textwidth} p{0.05\textwidth}}
\hline
 &  \multicolumn{1}{c}{Image} & \multicolumn{1}{c}{Disparity} & \multicolumn{1}{c}{Platform} & \multicolumn{3}{c}{FPGA} & \multicolumn{3}{c}{Throughput} \\
 &  \multicolumn{1}{c}{resolution} & \multicolumn{1}{c}{range} & \multicolumn{1}{c}{} & \multicolumn{3}{c}{resources} & \multicolumn{3}{c}{} \\
 &  & &  & \multicolumn{1}{c}{LUT} & \multicolumn{1}{c}{FF} & \multicolumn{1}{c}{BRAM} & \multicolumn{1}{c}{FPS} & \multicolumn{1}{c}{MDE/s} & \multicolumn{1}{c}{MDE/s/KLUT} \\
\hline
\cite{Hofmann2016} & \multicolumn{1}{c}{1920x1080} & \multicolumn{1}{c}{128} & \multicolumn{1}{c}{Virtex-7} & \multicolumn{1}{c}{195k} & \multicolumn{1}{c}{217k} & \multicolumn{1}{c}{368} & \multicolumn{1}{c}{30} & \multicolumn{1}{c}{7 963} &
\multicolumn{1}{c}{40.84} \\
\hline
\cite{Wang2015} & \multicolumn{1}{c}{1600x1200} & \multicolumn{1}{c}{128} & \multicolumn{1}{c}{Stratix-V} & \multicolumn{1}{c}{222k} & \multicolumn{1}{c}{149k} & \multicolumn{1}{c}{N/A} & \multicolumn{1}{c}{43} & \multicolumn{1}{c}{10 472}
& \multicolumn{1}{c}{47.2}\\
\hline
\cite{Shrivastava2020} & \multicolumn{1}{c}{1280x960} & \multicolumn{1}{c}{64} & \multicolumn{1}{c}{Virtex-7 690T} & \multicolumn{1}{c}{211k} & \multicolumn{1}{c}{N/A} & \multicolumn{1}{c}{641} & \multicolumn{1}{c}{322} & \multicolumn{1}{c}{25 056}  & \multicolumn{1}{c}{118.6} \\
\hline
\cite{Lee2021} & \multicolumn{1}{c}{1920x1080} & \multicolumn{1}{c}{128} & \multicolumn{1}{c}{Zynq US+} & \multicolumn{1}{c}{222k} & \multicolumn{1}{c}{135k} & \multicolumn{1}{c}{252} & \multicolumn{1}{c}{103} & \multicolumn{1}{c}{27 297} & \multicolumn{1}{c}{123.0} \\
\hline
\multicolumn{1}{c}{\textbf{New}} & \multicolumn{1}{c}{3840x2160} & \multicolumn{1}{c}{64} & \multicolumn{1}{c}{Virtex-7 485T} & \multicolumn{1}{c}{138k} & \multicolumn{1}{c}{65k} & \multicolumn{1}{c}{197} & \multicolumn{1}{c}{30} & \multicolumn{1}{c}{15 925} & \multicolumn{1}{c}{116.2}\\
\hline

& & & & & & & & \\
\end{tabular}
\end{table}






\section{Conclusion} 
\label{sec:conc}

In this paper, we presented a~hardware architecture for an SGM algorithm to process a~4K/Ultra HD video stream in real-time. 
We proposed a~solution to the inherent data dependency problem.
It allowed us to maintain high accuracy of the depth map estimation, while making it possible to take advantage of the 4ppc vector format.
We implemented the module on a~Virtex-7 FPGA platform achieving 30 frames per second for a~resolution of $3840 \times 2160$ pixels with 64 disparity levels.


In future work, we plan to add more aggregation paths to the algorithm. With that, it will be possible to get more accurate results, but at the cost of latency and resource usage. We also plan to implement a~video stream rectification module.

\subsubsection*{Acknowledgements} 
The work presented in this paper was supported by: the  National Science Centre project no. 2016/23/D/ST6/01389 entitled ''The development of computing resources organization in latest generation of heterogeneous reconfigurable devices enabling real-time processing of UHD/4K video stream'', the AGH University of Science and Technology project no. 16.16.120.773 and the program ''Excellence initiative –- research university'' for the AGH University of Science and Technology.

\printbibliography





\end{document}